\documentclass[11pt,a4paper]{article}
\usepackage[hyperref]{acl2017}
\usepackage{times}
\usepackage{latexsym}
\usepackage{graphicx}
\usepackage{amsfonts}
\usepackage{amsmath}

\usepackage{url}

\DeclareMathOperator*{\softmax}{softmax}

\DeclareMathOperator{\LSTM}{LSTM}

\graphicspath{ {images/} }

\title{Opinion Recommendation using Neural Memory Model}

\author{Zhongqing Wang and Yue Zhang\\
	   Singapore University of Technology and Design, \\
	    8 Somapah Road, Singapore 487372\\
	    {\tt \{zhongqing\_wang, yue\_zhang\}@sutd.edu.sg}}

\date{}

\begin{document}

\maketitle

\begin{abstract}
  We present opinion recommendation, a novel task of jointly predicting a custom review with a rating score that a certain user would give to a certain product or service, given existing reviews and rating scores to the product or service by other users, and the reviews that the user has given to other products and services.
  A characteristic of opinion recommendation is the reliance of multiple data sources for multi-task joint learning, which is the strength of neural models. We use a single neural network to model users and products, capturing their correlation and generating customised product representations using a deep memory network, from which customised ratings and reviews are constructed jointly. Results show that our opinion recommendation system gives ratings that are closer to real user ratings on Yelp.com data compared with Yelp's own ratings, and our methods give better results compared to several pipelines baselines using state-of-the-art sentiment rating and summarization systems.
\end{abstract}

\section{Introduction}

\begin{figure}[!tp]
\centering
\includegraphics[width=200px]{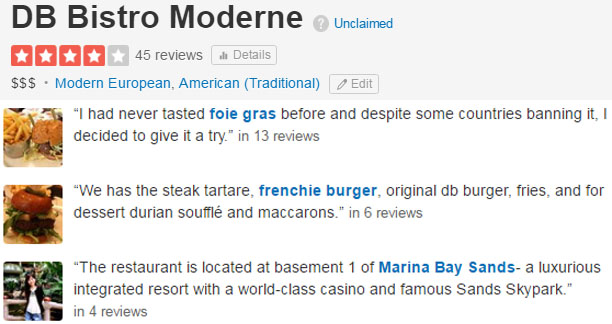}
\caption{\label{figure-example} A restaurant review on Yelp.com.}
\end{figure}

Offering a channel for customers to share opinions and give scores to products and services, review websites have become a highly influential information source that customers refer to for making purchase decisions. Popular examples include IMDB\footnote{\url{http://www.imdb.com/}} on the movie domain, Epinions\footnote{\url{http://epinions.com/}} on the product domain, and Yelp\footnote{\url{https://www.yelp.com/}} on the service domain. Figure~\ref{figure-example} shows a screenshot of a restaurant review page on Yelp.com, which offers two main types of information. First, an overall rating score is given under the restaurant name; second, detailed user reviews are listed below the rating.

Though offering a useful overview and details about a product or service, such information has several limitations. First, the overall rating is general and not necessarily agreeable to the taste of individual customers. Being a simple reflection of all customer scores, it serves an average customer well, but can be rather inaccurate for individuals. For example, the authors themselves often find highly rated movies being tedious. Second, there can be hundreds of reviews for a product or service, which makes it infeasible for exhaustive reading. It would be useful to have a brief summary of all reviews, which ideally should be customized to the reader.

We investigate the feasibility of a model that addresses the limitations above. There are two sources of information that the model should collect to achieve its goal, namely information on the \emph{target product}, and information about the \emph{user}. The former can be obtained from reviews written by other customers about the target product, and the latter can be obtained from the reviews that the user has written for other products and services. Given the above two sources of information, the model should generate a \emph{customized score} of the product that the user is likely to give after trying, as well as a \emph{customized review} that the user would have written for the target product.

We refer to the task above using the term \textbf{opinion recommendation}, which is a new task, yet closely related to several existing lines of work in NLP. The first is \emph{sentiment analysis}~\cite{HuL04,PangL07}, which is to give a rating score based on a customer review. Our task is different in that we aim to predict user rating scores of new product, instead of predicting the opinion score of existing reviews. The second is \emph{opinion summarization}~\cite{NishikawaHMK10,WangL16}, which is to generate a summary based on reviews of a product. A major difference between our task and this task is that the summary must be customized to a certain user, and a rating score must additionally be given. The third is \emph{recommendation}~\cite{su2009survey,yang2014survey}, which is to give a ranking score for a certain product or service based on the purchase history of the user and other customers who have purchased the target product. Our task is different in the source of input, which is \emph{textual} customer reviews and ratings rather than \emph{numerical} purchase history.

There are three types of inputs for our task, namely the reviews of the target product, the reviews of the user on other products, and other users reviews on other products, and two types of outputs, namely a customized rating score and a customized review. The ideal solution should consider the interaction between all given types of information, jointly predicting the two types of outputs. This poses significant challenges to statistical models, which require manually defined features to capture relevant patterns from training data. Deep learning is a relatively more feasible choice, offering viabilities of information fusion by fully connected hidden layers~\cite{CollobertWBKKK11,HendersonMTM13}. We leverage this advantage in building our model.

In particular, we use a recurrent neural network to model the semantic content of each review. A neural network is used to consolidate existing reviews for the target product, serving the role of a \emph{product model}. In addition, a \emph{user model} is built by consolidating the reviews of the given user into a single vector form. Third, to address potential sparsity of a user's history reviews, neighbor users are identified by collaborative filtering~\cite{ding2006orthogonal}, and a vector representation is learned by using a neural \emph{neighborhood model}, which consolidates their history reviews. Finally, a deep memory network is utilized to find the association between the user and target product, jointly yielding the rating score and customised review.

Experiments on a Yelp dataset show that the model outperforms several pipelined baselines using state-of-the-art techniques. In particular, review scores given by the opinion recordation system are closer to real user review scores compared to the review scores which Yelp assigns to target products. Our code is released at \url{http://github.com/anonymous}.

\section{Related Work}

\textbf{Sentiment Analysis.} Our task is related to document-level sentiment classification~\cite{PangL07}, which is to infer the sentiment polarity of a given document. Recently, various neural network models are used to capture the sentimental information automatically, including convolutional neural networks~\cite{Kim14}, recursive neural network~\cite{socher2013recursive} and recurrent neural network~\cite{TengVZ16,TaiSM15}, which have been shown to achieve competitive results across different benchmarks. Different from binary classification, review rating prediction aims to predict the numeric rating of a given review. \newcite{PangL05} pioneered this task by regarding it as a classification/regression problem. Most subsequent work focuses on designing effective textural features of reviews~\cite{QuIW10,LiLJZYZ11,Wan13}. Recently, \newcite{TangQLY15} proposed a neural network model to predict the rating score by using both lexical semantic and user model.

Beyond textural features, user information is also investigated in the literature of sentiment analysis. For example, \newcite{gao2013modeling} developed user-specific features to capture user leniency, and \newcite{li2014suit} incorporated textual topic and user-word factors through topic modeling. For integrating user information into neural network models, \newcite{TangQLY15} predicted the rating score given a review by using both lexical semantic information and a user embedding model. \newcite{ChenSTLL16} proposed a neural network to incorporate global user and product information for sentiment classification via an attention mechanism.

Different from the above research on sentiment analysis, which focuses on predicting the opinion on existing reviews. Our task is to recommend the score that a user would give to a new product \emph{without} knowing his review text. The difference originates from the object, previous research aims to predict opinions on reviewed products, while our task is to recommend opinion on new products, which the user has not reviewed.

\textbf{Opinion Summarization.} Our work also overlaps with to the area of opinion summarization, which constructs natural language summaries for multiple product reviews~\cite{HuL04}. Most previous work extracts opinion words and aspect terms. Typical approaches include association mining of frequent candidate aspects~\cite{HuL04,QiuLBC11}, sequence labeling based methods~\cite{JakobG10,YangC13}, as well as topic modeling techniques~\cite{LinH09}. Recently, word embeddings and recurrent neural networks are also used to extract aspect terms~\cite{irsoy2014opinion,LiuJM15}.

Aspect term extraction approaches lack critical information for a user to understand how an aspect receives a particular rating. To address this, \newcite{NishikawaHMK10} generated summaries by selecting and ordering sentences taken from multiple review texts according to affirmativeness and readability of the sentence order. \newcite{WangL11} adopted both sentence-ranking and graph-based methods to extract summaries on an opinion conversation dataset.
While all the methods above are \emph{extractive}, \newcite{ganesan2010opinosis} presented a graph-based summarization framework to generate concise abstractive summaries of highly redundant opinions, and \newcite{WangL16} used an attention-based neural network model to absorb information from multiple text units and generate summaries of movie reviews.

Different from the above research on opinion summarization, we generate a \emph{customized} review to a certain user, and a rating score must be additionally given.

\textbf{Recommendation.} Recommendation systems suggest to a user new products and services that might be of their interest. There are two main approaches, which are content-based and collaborative-filtering (CF) based~\cite{adomavicius2005toward,yang2014survey}, respectively. Most existing social recommendation systems are CF-based, and can be further grouped into model-based CF and neighborhood-based CF~\cite{kantor2011recommender,su2009survey}. Matrix Factorization (MF) is one of the most popular models for CF. In recent MF-based social recommendation works, user-user social trust information is integrated with user-item feedback history (e.g., ratings, clicks, purchases) to improve the accuracy of traditional recommendation systems, which only factorize user-item feedback data~\cite{ding2006orthogonal,koren2008factorization,HeZKC16}.

There has been work integrating sentiment analysis and recommendation systems, which use recommendation strategies such as matrix factorization to improve the performance of sentiment analysis~\cite{leung2006integrating,SinghMM11}. These methods typically use ensemble learning~\cite{SinghMM11} or probabilistic graph models~\cite{WuE15}. For example, \newcite{ZhangL0ZLM14} who proposed a factor graph model to recommend opinion rating scores by using explicit product features as hidden variables.

Different from the above research on recommendation systems, which utilize numerical purchase history between users and products, we work with textual information. In addition, recommendation systems only predict a rating score, while our system generates also a customized review, which is more informative.

\textbf{Neural Network Models.} Multi-task learning has been recognised as a strength of neural network models for natural language processing~\cite{CollobertWBKKK11,HendersonMTM13,ZhangW16,ChenZL16}, where hidden feature layers are shared between different tasks that have common basis. Our work can be regarded as an instance of such multi-tasks learning via shared parameters, which has been widely used in the research community recently.

Dynamic memory network models are inspired by neural turing machines~\cite{GravesWD14}, and have been applied for NLP tasks such as question answering~\cite{SukhbaatarSWF15,KumarIOIBGZPS16}, language modeling~\cite{TranBM16} and machine translation~\cite{WangLLL16}. It is typically used to find abstract semantic representations of texts towards certain tasks, which are consistent with our main need, namely abstracting the representation of a product that is biased towards the taste of a certain user.

\section{Model}

\begin{figure}[!tp]
\setlength{\abovecaptionskip}{0cm}
\centering
\includegraphics[width=200px]{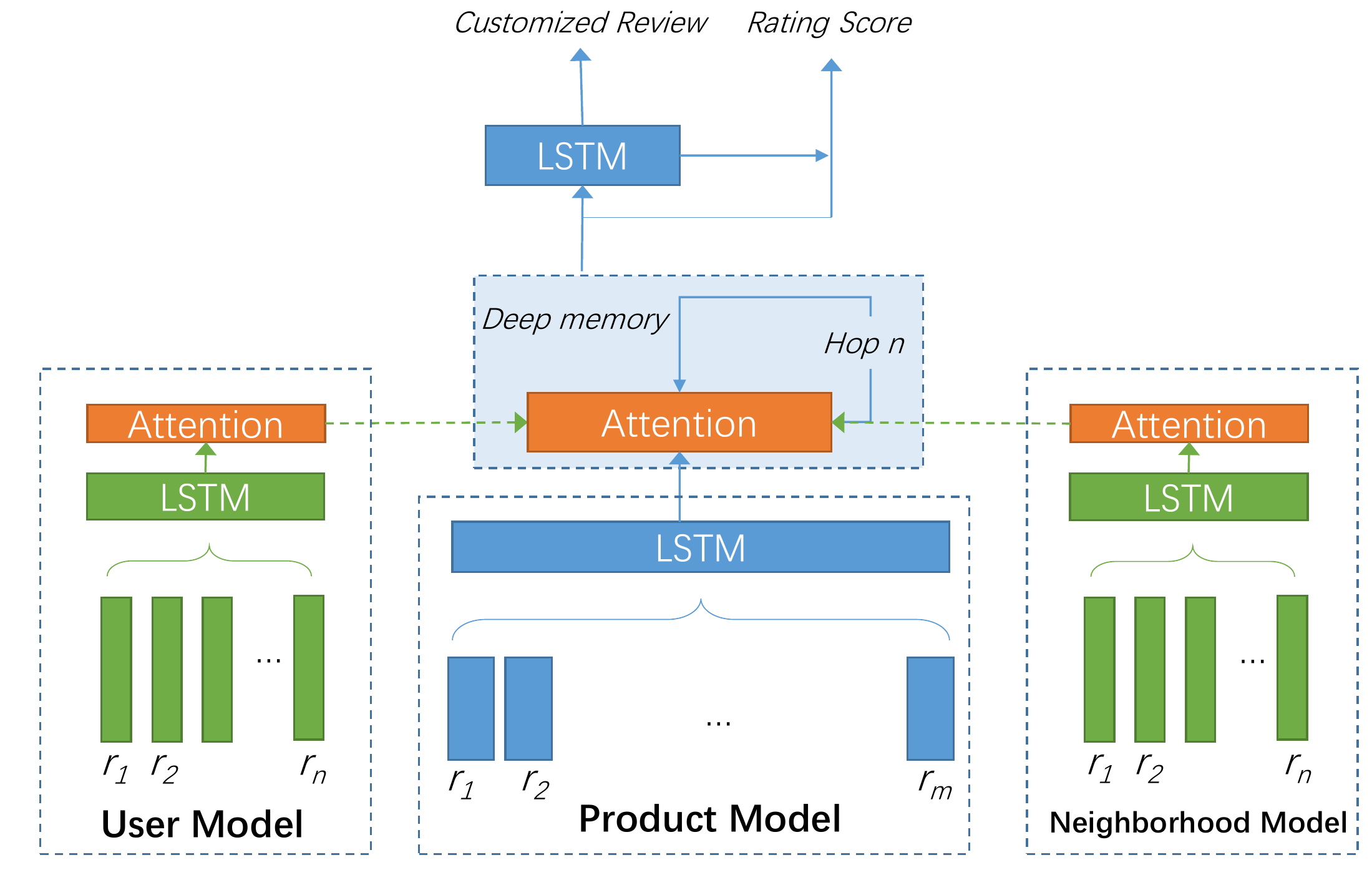}
\caption{\label{model} Overview of proposed model.}
\end{figure}

Formally, the input to our model is a tuple $\langle R_T, R_U, R_N \rangle$, where $R_T = \{r_{T_1},r_{T_2},...,r_{T_{n_t}}\}$ is the set of existing reviews of a target product, $R_U=\{r_{U_1},r_{U_2},...,r_{U_{n_u}}\}$ is the set of user's history reviews, and $R_N=\{r_{N_1},r_{N_2},...,r_{N_{n_n}}\}$ is the set of the user's neighborhood reviews. All the reviews are sorted with temporal order. The output is a pair $\langle Y_S,Y_R \rangle$, where $Y_S$ is a real number between 0 and 5 representing the rating score of the target product, and $Y_R$ is a customised review.

For capturing both general and personalized information, we first build a \emph{product model}, a \emph{user model}, and a \emph{neighborhood model}, respectively, and then use a memory network model to integrate these three types of information, constructing a \emph{customized product model}. Finally, we predict a customized rating score and a review collectively using neural stacking. The overall architecture of the model is shown in Figure~\ref{model}.

\subsection{Review Model}

A customer review is the foundation of our model, based on which we derive representations of both a user and a target product. In particular, a user profile can be achieved by modeling all the reviews of the user $R_U$, and a target product profile can be obtained by using all existing reviews of the product $R_T$. We use the average of word embeddings to model a review. Formally, given a review $r=\{x_1,x_2,...,x_m\}$, where $m$ is the length of the review, each word $x_k$ is represented with a \emph{K}-dimensional embedding $e_k^w$~\cite{MikolovSCCD13}. We use the $\sum_k(e_k^w)/m$ for the representation of the review $e_{r}^d$.

\subsection{User Model}\label{section-user-model}

A standard LSTM~\cite{HochreiterS97} without coupled input and forget gates or peephole connections is used to learn the hidden states of the reviews. Denoting the recurrent function at step $t$ as $\LSTM(x_t,h_{t-1})$, we obtain a sequence of hidden state vectors $\{h_{U_1},h_{U_2},...,h_{U_{n_u}}\}$ recurrently by feeding $\{e^d(r_{U_1}),e^d(r_{U_2}),...,e^d{r_{U_{n_u}}}\}$ as inputs, where $h_{U_i}=\LSTM(e^d(r_{U_i}),h_{U_{i-1}})$. The initial state and all stand LSTM parameters are randomly initialized and tuned during training.

Not all reviews contribute equally to the representation of a user. We introduce an attention mechanism~\cite{BahdanauCB14,Yang16} to extract the reviews that are relatively more important, and aggregate the representation of reviews to form a vector. Taking the hidden state $\{h_{U_1},...h_{U_2},...,h_{U_{n_u}}\}$ of user model as input, the attention model outputs, a continuous vector $v_U \in \mathbb{R}^{d \times 1}$, which is computed as a weighted sum of each hidden state $h_{U_i}$, namely
\begin{equation} \label{eq-9}
v_U=\sum_{i}^{n_u}\alpha_{i}h_{U_i}
\end{equation}
where $n_u$ is the hidden variable size, $\alpha_i \in [0,1]$ is the weight of $h_{U_i}$, and $\sum_i\alpha_i=1$.

For each piece of hidden state $h_{U_i}$, the scoring function is calculated by
\begin{equation} \label{eq-7}
u_{i}=\tanh(W_Uh_{U_i}+b_U)
\end{equation}
\begin{equation} \label{eq-8}
\alpha_{i}=\frac{\exp(u_i)}{\sum_{j}{\exp(u_j)}}
\end{equation}
where $W_U$ and $b_U$ are model parameters. The attention vector $v_U$ is used to represent the \emph{User Model}.

\subsection{Finding Neighbor Users} \label{section-neighbor}

We use neighborhood reviews to improve the user model, since a user may not have sufficient reviews to construct a reliable model. Here a neighbor refers to a user that has similar tastes to the target user~\cite{koren2008factorization,DesrosiersK11}. The same as the user model, we construct the \emph{neighborhood model} $v_N$ using the neighborhood reviews $R_N=\{r_{N_1},r_{N_2},...,r_{N_{n_n}}\}$ with an attention recurrent network.

A key issue in building the neighborhood model is how to find neighbors of a certain user. In this study, we use matrix factorization~\cite{koren2008factorization} to detect neighbors, which is a standard approach for recommendation~\cite{ding2006orthogonal,li2009non,HeZKC16}. In particular, users' rating scores of products are used to build a product-users matrix $M \in \mathbb{R}^{n_t \times n_u}$ with $n_t$ products and $n_u$ users. We approximate it using three factors, which specify soft membership of products and users~\cite{ding2006orthogonal} by finding:
\begin{equation} \label{eq-mf}
\begin{split}
\min_{F,S,T}||M-FST^T||\\ s.t. S\geq0,F\geq0,T\geq0
\end{split}
\end{equation}
where  $F\in \mathbb{R}^{n_t \times K}$ represents the posterior probability of $K$ topic clusters for each product; $S\in \mathbb{R}^{K \times K}$ encodes the distribution of each topic $k$; and $T\in \mathbb{R}^{K \times n_u}$ indicates the posterior probability of $K$ topic clusters for each user.

As a result of matrix factorization, we directly obtain the probability of each user on each topic from the person-topic matrix $T$. To infer $T$, the optimization problem in Eq.\ref{eq-mf} can be solved using the following updating rule:
\begin{equation}
T_{jk}\leftarrow T_{jk}\frac{(M^TFS)_{jk}}{(TT^TM^TFS)_{jk}}
\end{equation}
Obtaining the user-topic matrix $T$, we measure the implicit connection between two users using:
\begin{equation}
sim(i,j)=\sum^{k}_{k=1}T_{ik}T_{jk}
\end{equation}
where $sim(i,j)$ measure the implicit connection degree between users $i$ and $j$. If $sim(i,j)$ is higher than a threshold $\eta$, we consider user $j$ as the neighbor of user $i$.

\subsection{Product Model} \label{section-target-model}

Given the representations of existing reviews $\{e(r_{T_1}),e(r_{T_2}),...,e{r_{T_{n_t}}}\}$ of the product, we use a LSTM to model their temporal orders, obtaining a sequence of hidden state vectors $h_T=\{h_{T_1},h_{T_2},...,h_{T_{n_t}}\}$ by recurrently feeding $\{e(r_{T_1}),e(r_{T_2}),...,e{r_{T_{n_t}}}\}$ as inputs. The hidden state vectors $h_T$ are used to represent the product.

\subsection{Customized Product Model}

We use the user representation $v_U$ and the neighbour representation $v_N$ to transform the target product representation $h_T=\{h_{T_1},h_{T_2},...,h_{T_{n_t}}\}$ into a customised product representation $v_C$, which is tailored to the taste of the user. In particular, a dynamic memory network~\cite{SukhbaatarSWF15,XiongMS16} is utilized to iteratively find increasingly abstract representations of $h_t$, by injecting $v_U$ and $v_N$ information.

The memory model consists of multiple dynamic computational layers (hops), each of which contains an attention layer and a linear layer. In the first computational layer (hop 1), we take the hidden variables $h_{T_i}$ ($0\leq i\leq n_t$) of product model as input, adaptively selecting important evidences through one attention layer using $v_U$ and $v_N$. The output of the attention layer gives a linear interpolation of $h_T$, and the result is considered as input to the next layer (hop 2). In the same way, we stack multiple hops and run the steps multiple times, so that more abstract representations of the target product can be derived.

The attention model outputs a continuous vector $v_C \in \mathbb{R}^{d \times 1}$, which is computed as a weighted sum of $h_{T_i}$ ($0\leq i\leq n_t$), namely
\begin{equation} \label{eq-9}
v_C=\sum_{i}^{n_t}\beta_{i}h_{T_i}
\end{equation}
where $n_t$ is the hidden variable size, $\beta_i \in [0,1]$ is the weight of $h_{T_i}$, and $\sum_i\beta_i=1$. For each piece of hidden state $h_{T_i}$, we use a feed forward neural network to compute its semantic relatedness with the abstract representation $v_C$. The scoring function is calculated as follows at hop $t$:
\begin{equation} \label{eq-7}
\begin{split}
u_{i}^t=\tanh(W_Th_{T_i}+W_Cv_C^{t-1}\\+W_Uv_U+W_Nv_N+b)
\end{split}
\end{equation}
\begin{equation} \label{eq-8}
\beta_{i}^t=\frac{\exp(u_i^t)}{\sum_{j}{\exp(u_j^t)}}
\end{equation}
The vector $v_C$ is used to represent the customized product model. At the first hop, we define $V_C^0=\sum_nh_{T_i}/n$.

The product model  $h_{T_i}$ ($0\leq i\leq n_t$) represents salient information of existing reviews in their temporal order, they do not reflect the taste of a particular user. We use the customised product model to integrate user information and  product information (as reflected by the product model), resulting in a single vector that represents a customised product. From this vector we are able to synthesis both a customised review and a customised rating score.

\subsection{Customized Review Generation}

The goal of customized review generation is to generate a review $Y_R$ from the customized product representation $v_C$, composed by a sequence of words $y_{R_1},...,y_{R_{n_r}}$. We decompose the prediction of $Y_R$ into a sequence of word-level predictions:
\begin{equation} \label{eq-summary}
\begin{split}
\log &P(Y_R|v_C)=\\
&\sum_{j}P(y_{R_j}|y_{R_1},...,y_{R_{j-1}},v_C)
\end{split}
\end{equation}
where each word $y_{R_j}$ is predicted conditional on the previously generated $y_{R_1},...,y_{R_{j-1}}$ and the input $v_C$. The probability is estimated by using standard word softmax:
\begin{equation}
\begin{split}
P(y_{R_j}|y_{R_1},...,y_{R_{j-1}},v_C)&=\\
&\softmax(h_{R_j})
\end{split}
\end{equation}
where $h_{R_j}$ is the hidden state variable at timestamp $j$, which is modeled as $LSTM(u_{j-1},h_{Rj})$.
Here a LSTM is used to generate a new state $h_{R_j}$ from the representation of the previous state $h_{R_{j-1}}$ and $u_{j-1}$. $u_{j-1}$ is the concatenation of previously generated word $y_{R_{j-1}}$ and the input representation of customized model $v_C$.

\subsection{Customized Opinion Rating Prediction}

We consider two factors for customised opinion rating, namely existing review scores and the customised product representation $v_C$. A baseline rating system such as Yelp.com uses only the former information, typically by taking the average of existing review scores. Such a baseline gives an empirical square error of 1.28 (out of 5) in our experiments, when compared with a test set of individual user ratings, which reflects the variance in user tastes. In order to integrate user preferences into the rating, we instead take a weighted average of existing ratings cores, so that the scores of reviews that are closer to the user preference are given higher weights.

As a second factor, we calculate a review score independently according to the customised representation $v_c$ of existing reviews, without considering review scores. The motivation is two fold. First, existing reviews can be relatively few, and hence using their scores alone might not be sufficient for a confident score. Second, existing ratings can be all different from a user¡¯s personal rating, if the existing reviews do not come from the user's neighbours. As a result, using the average or weighted average of existing reviews, the personalised user rating might not be reached.

Formally, given the rating scores $s_1,s_2,...,s_n$ of existing reviews, and the the customized product representation $v_C$, we calculate:
\begin{equation}
Y_S=\sum_i^n\alpha_i \cdot s_i + \mu\tanh(W_Sv_C+b_S)
\end{equation}
In the left term $\sum_i^n\alpha_i \cdot s_i$, we use attention weights $\alpha_i$ to measure the important of each rating score $s_i$. The right term $\tanh(W_Sv_C+b_S)$ is a review-based shift, weighted by $\mu$.

Since the result of customized review generation can be helpful for rating score prediction, we use neural stacking additionally feeding the last hidden state $h_{R_n}$ of review generation model as input for $Y_S$ prediction, resulting in
\begin{equation} \label{eq-stacking}
\begin{split}
Y_S=&\sum_i^n\alpha_i \cdot s_i + \\
&+\mu\tanh(W_S(v_C \oplus h_{R_n})+b_S)
\end{split}
\end{equation}
where $\oplus$ denotes vector concatenation.

\subsection{Training}
For our task, there are two joint training objectives, for review scoring and review summarisation, respectively. The loss function for the former is defined as:
\begin{equation}
L(\Theta)=\sum_{i=1}^N(Y_{S_i}^*-Y_{S_i})^2+\frac{\lambda}{2}||\Theta||^2
\end{equation}
where $Y_{S_i}^*$ is the predicted rating score, $Y_{S_i}$ is the rating score in the training data, $\Theta$ is the set of model parameters and $\lambda$ is a parameter for L2 regularization.

We train the customized review generation model by maximizing the log probability of Eq.\ref{eq-summary}~\cite{SutskeverVL14,RushCW15}. Standard back propagation is performed to optimize parameters, where gradients also propagate from the scoring objective to the review generation objective due to neural stacking (Eq.\ref{eq-stacking}). We apply online training, where model parameters are optimized by using Adagrad~\cite{DuchiHS11}. For all LSTM models, we empirically set the size of the hidden layers to 128. We train word embeddings using the \emph{Skip-gram} algorithm~\cite{MikolovSCCD13}\footnote{ https://code.google.com/p/word2vec/}, using a window size of 5 and vector size of 128. In order to avoid over-fitting, dropout~\cite{Hinton12} is used for word embedding with a ratio of 0.2. The neighbor similarity threshold $\eta$ is set to 0.25.

\section{Experiments}

\subsection{Experimental Settings}

Our data are collected from the yelp academic dataset\footnote{https://www.yelp.com/academic\_dataset}, provided by Yelp.com, a popular restaurant review website. The data set contains three types of objects: \emph{business}, \emph{user}, and \emph{review}, where business objects contain basic information about local businesses (i.e. restaurants), review objects contain review texts and star rating, and user objects contain aggregate information about a single user across all of Yelp. Table~\ref{table-data} illustrates the general statistics of the dataset.

\begin{table}
\centering
\begin{tabular}{l|c}
\hline   & \bf Amount \\ \hline \hline
Business & 15,584 \\
Review & 334,997 \\
User &  303,032\\
\hline
\end{tabular}
\caption{\label{table-data} Statistics of the dataset.}
\end{table}

For evaluating our model, we choose 4,755 user-product pairs from the dataset. For each pair, the existing reviews of the target service (restaurant) are used for the product model. The rating score given by each user to the target service is considered as the gold customized rating score,  and the review of the target service given by each user is used as the gold-standard customized review for the user. The remaining reviews of each user are used for training the user model. We use 3,000 user-product pairs to train the model, 1,000 pairs as testing data, and remaining data for development.

We use the ROUGE-1.5.5~\cite{lin2004rouge} toolkit for evaluating the performance of customized review generation, and report unigram overlap (ROUGE-1) as a means of assessing informativeness. We use Mean Square Error (MSE)~\cite{Wan13,TangQLY15} is used as the evaluation metric for measuring the performance of customized rating score prediction. MSE penalizes more severe errors more heavily.

\subsection{Development Experiments}

\subsubsection{Ablation Test}

Effects of various configurations of our model, are shown on Table~\ref{table-factors}, where \emph{Joint} is the full model of this paper, \emph{-user} ablates the user model, \emph{-neighbor} ablates the neighbor model, \emph{-rating} is a single-task model that generates a review without the rating score, and \emph{-generation} generates only the rating score.

By comparing ``Joint'' and ``-user,-neighbor'', we can find that customized information have significant influence on both the rating and review generation results ($p-value<0.01$ using $t$-test). In addition, comparison between ``-Joint'' and ``-user'', and between ``-user'' and ``-user, -neighbor'' shows that both the user information and the neighbour user information of the user are effective for improving the results. A user¡¯s neighbours can indeed alleviate scarcity of user reviews.

Finally, comparison between ``Joint'' and ``-generation'', and between ``Joint'' and ``-rating'' shows that multi-task learning by parameter sharing is highly useful.

\begin{table}
\centering
\begin{tabular}{l|c|c}
\hline   & \bf Rating & \bf Generation \\ \hline \hline
Joint & 0.904 & 0.267 \\ \hline
-user & 1.254 & 0.220 \\
-neighbor & 1.162 & 0.245 \\
-user,-neighbor & 1.342 & 0.205 \\
-rating & -  & 0.254 \\
-generation & 1.042 & - \\
\hline
\end{tabular}
\caption{\label{table-factors} Feature ablation tests.}
\end{table}

\subsubsection{Influence of Hops}

\begin{figure}[!tp]
\setlength{\abovecaptionskip}{0cm}
\centering
\includegraphics[width=200px]{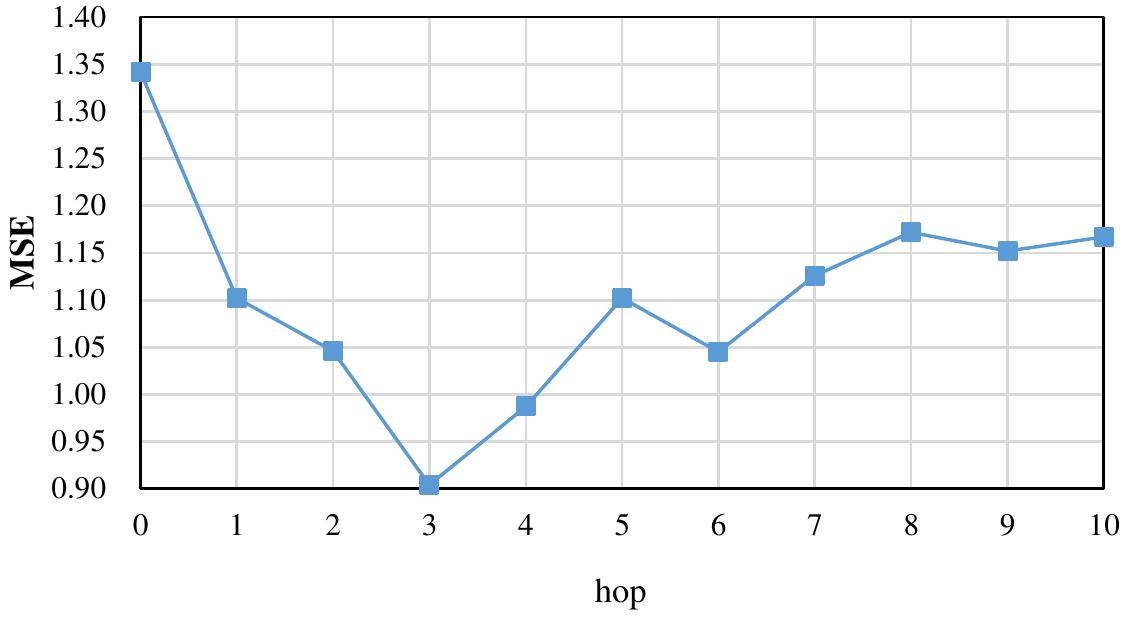}
\caption{\label{figure-hop} Influence of hops.}
\end{figure}

We show the influence of hops of memory network for rating prediction on Figure~\ref{figure-hop}. Note that, the model would only consider the general product reviews ($-user,-neighbor$), when $hop=0$. From the figure we can find that, when $hop=3$, the performance is the best. It indicates that multiple hops can capture more abstract evidences from external memory to improve the performance. However, too many hops leads to over-fitting, thereby harms the performance. As a result, we choose 3 as the number of hops in our final test.

\subsubsection{Influence of $\mu$}

\begin{figure}[!tp]
\setlength{\abovecaptionskip}{0cm}
\centering
\includegraphics[width=200px]{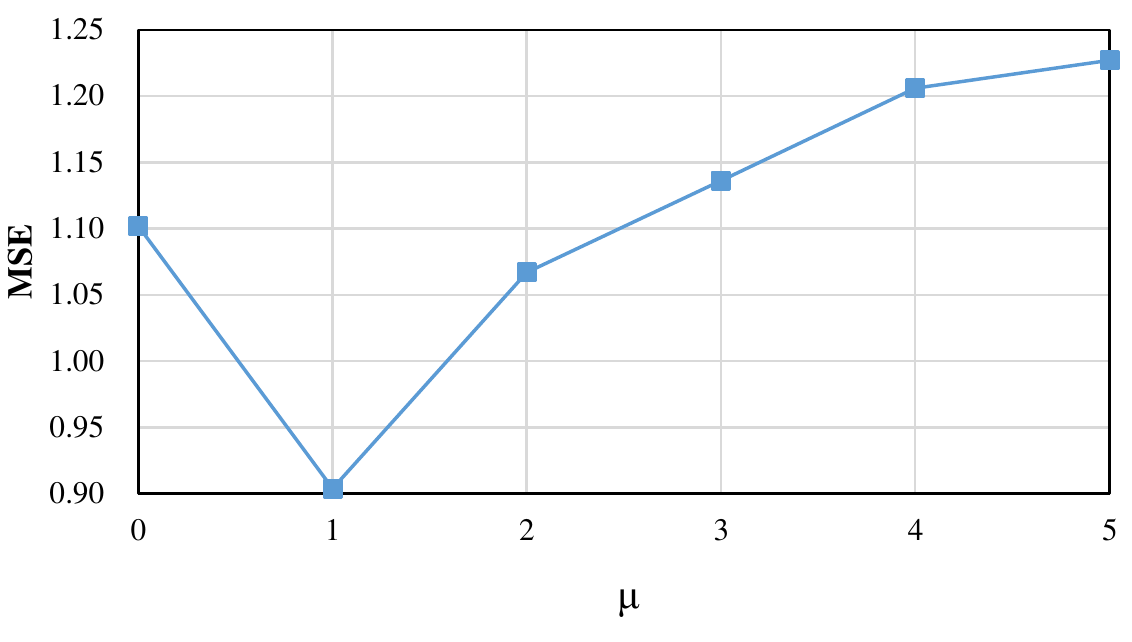}
\caption{\label{figure-bias} Influence of bias score.}
\end{figure}

We show the influence of the bias weight parameter $\mu$ for rating prediction in Figure~\ref{figure-bias}. With $\mu$ being 0, the model uses the weighted sum of existing reviews to score the product. When $\mu$ is very large, the system tends to use only the customized product representation $v_c$ to score the product, hence ignoring existing review scores, which are a useful source of information. Our results show that when $\mu$ is 1, the performance is optimal, thus indicating both existing review scores and review contents are equally useful.

\subsection{Final Results}

We show the final results for opinion recommendation, comparing our proposed model with the following state-of-the-art baseline systems:

\begin{itemize}
\item \emph{RS-Average} is the widely-adopted baseline (e.g., by Yelp.com), using the averaged review scores as the final score.

\item \emph{RS-Linear} estimates the rating score that a user would give by $s_{ui}=s_{all}+s_{u}+s_{i}$~\cite{Ricci2011}, where $s_u$ and $s_i$ are the the training deviations of the user $u$ and the product $i$, respectively.

\item \emph{RS-Item} applies $k$NN to estimate the rating score~\cite{SarwarKKR01}. We choose the cosine similarity between $v_c$ to measure the distance between product.

\item \emph{RS-MF} is a state-of-the-art recommendation model, which uses matrix factorisation to predict rating score~\cite{ding2006orthogonal,li2009non,HeZKC16}.


\item \emph{Sum-Opinosis} uses a graph-based framework to generate abstractive summarisation given redundant opinions~\cite{ganesan2010opinosis}.

\item \emph{Sum-LSTM-Att} is a state-of-the-art neural abstractive summariser, which uses an attentional neural model to consolidate information from multiple text sources, generating summaries using LSTM decoding~\cite{RushCW15,WangL16}.
\end{itemize}

All the baseline models are single-task models, without considering rating and summarisation prediction jointly. The results are shown in Table~\ref{table-baseline}. Our model (`` Joint'') significantly outperforms both ``RS-Average'' and ``RS-Linear'' ($p-value<0.01$ using $t$-test), which demonstrates the strength of opinion recommendation, which leverages user characteristics for calculating a rating score for the user.

Our proposed model also significantly outperforms state-of-the-art recommendation systems (RS-Item and RS-MF) ($p-value<0.01$ using $t$-test), indicating that textual information are a useful addition to the rating scores themselves for recommending a product.

Finally, comparison between our proposed model and state-of-the-art summarisation techniques (Sum-Opinosis and Sum-LSTM-Att) shows the advantage of leveraging user information to enhance customised review generation, and also the strength of joint learning.

\begin{table}
\centering
\begin{tabular}{l|c|c}
\hline   & \bf Rating & \bf Generation \\ \hline \hline
RS-Average & 1.280 & - \\
RS-Linear & 1.234 & - \\
RS-Item & 1.364 & - \\
RS-MF & 1.143 & - \\
Sum-Opinosis & - & 0.183 \\
Sum-LSTM-Att & - & 0.196 \\ \hline
Joint & \bf 1.023 & \bf 0.250 \\
\hline
\end{tabular}
\caption{\label{table-baseline} Final results.}
\end{table}

\section{Conclusion}

We presented a dynamic memory model for opinion recommendation, a novel task of jointly predicting the review and rating score that a certain user would give to a certain product or service. In particular, a deep memory network was utilized to find the association between the user and the product, jointly yielding the rating score and customised review. Results show that our methods are better results compared to several pipelines baselines using state-of-the-art sentiment rating and summarisation systems.

\bibliography{acl2016}
\bibliographystyle{acl_natbib}

\end{document}